\documentclass[sigconf,trxiaoreview]{acmart}
\usepackage{algorithm}
\usepackage{algorithmicx}
\usepackage{algpseudocode}
\usepackage{comment}
\usepackage{amsmath}
\usepackage{multirow}
\usepackage{url}            
\usepackage{booktabs}       
\usepackage{amsfonts}       
\usepackage{nicefrac}       
\usepackage{microtype}
\usepackage{enumitem}
\usepackage[most]{tcolorbox}
\tcbuselibrary{listingsutf8,skins,breakable,most} 

\usepackage{subcaption}

\newcommand{\ours}{\texttt{Ensemble-RAG}}
\newcommand{\ie}{\textit{i.e.}}
\newcommand{\eg}{\textit{e.g.}}

\AtBeginDocument{%
  }

\setcopyright{acmlicensed}
\copyrightyear{2025}
\acmYear{2025}
\acmDOI{XXXXXXX.XXXXXXX}
\acmConference{CIKM'25}{November 10-14, 2025}{Seoul, Korea}
\acmISBN{978-1-4503-XXXX-X/2018/06}

\begin{document}


\title{Revisiting RAG Ensemble: A Theoretical and Mechanistic Analysis of Multi-RAG System Collaboration} 


\author{Yifei Chen}
\email{zhangboguodong@ruc.edu.cn}
\orcid{0009-0008-3142-5884}
\affiliation{%
  \institution{Gaoling School of Artificial Intelligence, Renmin University of China}
  \city{Haidian Qu}
  \state{Beijing Shi}
  \country{China}
}

\author{Guanting Dong}
\affiliation{%
  \institution{Gaoling School of Artificial Intelligence, Renmin University of China}
  \city{Haidian Qu}
  \state{Beijing Shi}
  \country{China}
}

\author{Yutao Zhu}
\affiliation{%
  \institution{Gaoling School of Artificial Intelligence, Renmin University of China}
  \city{Haidian Qu}
  \state{Beijing Shi}
  \country{China}
}

\author{Zhicheng Dou}
\affiliation{%
  \institution{Gaoling School of Artificial Intelligence, Renmin University of China}
  \city{Haidian Qu}
  \state{Beijing Shi}
  \country{China}
}

\renewcommand{\shortauthors}{Chen et al.}

\begin{abstract}

Retrieval-Augmented Generation (RAG) technology has been widely applied in recent years. However, despite the emergence of various RAG frameworks, a single RAG framework still cannot adapt well to a broad range of downstream tasks. Therefore, how to leverage the advantages of multiple RAG systems has become an area worth exploring. To address this issue, we have conducted a comprehensive and systematic investigation into ensemble methods based on RAG systems. Specifically, we have analyzed the RAG ensemble framework from both theoretical and mechanistic analysis perspectives. From the theoretical analysis, we provide the first explanation of the RAG ensemble framework from the perspective of information entropy. In terms of mechanism analysis, we have explored the RAG ensemble framework from both the pipeline and module levels. We carefully select four different pipelines (Branching, Iterative, Loop, and Agentic) and three different modules (Generator, Retriever, and Reranker) to solve seven different research questions. The experiments show that aggregating multiple RAG systems is both generalizable and robust, whether at the pipeline level or the module level. Our work lays the foundation for similar research on the multi-RAG system ensemble.
\end{abstract}





\maketitle



\section{Introduction}
\label{sec:introduction}
The emergence of Large Language Models (LLMs) has profoundly revolutionized many real-world tasks that rely on natural language~\citep{instructgpt,gpt3,llmsurvey}. However, when dealing with knowledge-intensive tasks, LLMs relying solely on their parametric knowledge often suffer from factual inconsistencies or hallucinations. To address these limitations, Retrieval-Augmented Generation (RAG) methods have been proposed, augmenting LLMs with dynamically retrieved external knowledge. This integration enhances response accuracy and reliability by grounding outputs in verifiable information sources.

As research in this field advances, more and more RAG methods have been proposed. Component Module RAG inserts various modules into the standard pipeline to better complete the retrieval task. For instance, the LongLLMLingua and RECOMP methods refine the retrieved knowledge with a refiner, and the SKR and Adaptive RAG methods distinguish the difficulty of questions by introducing a Judger~\citep{longllmlingua,RECOMP,SKR,adaptiverag}. Pipeline Module RAG optimizes the whole process to improve the accuracy and efficiency. For example, RePlug method is suitable for tasks with varying difficulty levels, and methods such as Iter-RetGen and Self-RAG are suitable for solving multi-hop problems~\citep{replug,modularRAG2,SELF-RAG}. With the development of agent technology, the application of Agentic RAG technology is becoming increasingly widespread~\citep{webshaper,webdancer,webwalker,websailor}. For example, Search-o1 and WebThinker combine search and reasoning and perform well in deep search tasks~\citep{Search_o1,webthinker}. However, given the inherent complexity of tasks and the heterogeneity of RAG workflows, developing a universal RAG framework that generalizes effectively across diverse applications remains a significant challenge.

\begin{figure}[t]
    \centering
    \includegraphics[width=\linewidth]{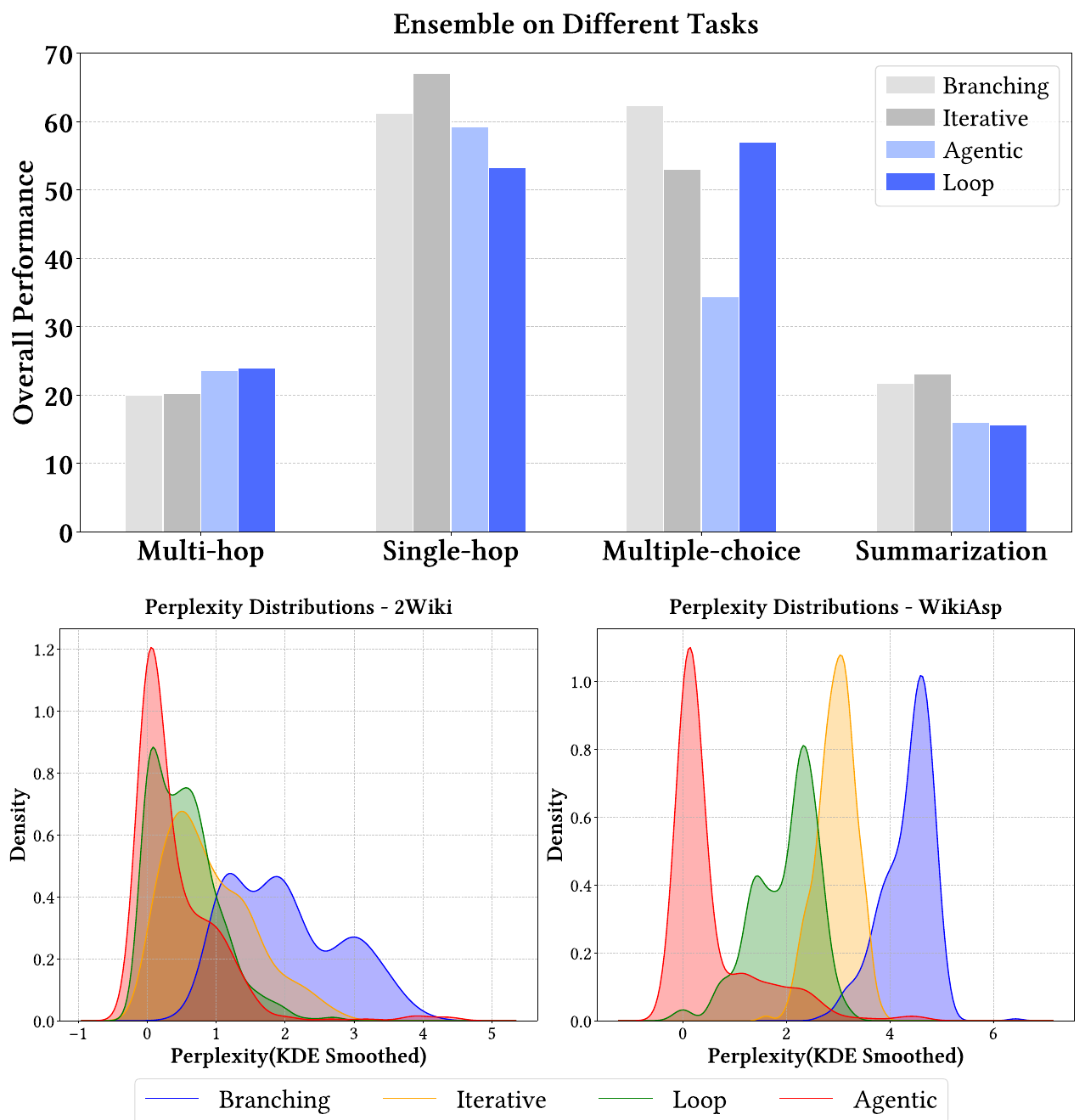}
    \caption{Overall Performance of Base RAG and Ensemble Methods. \textbf{Top:} the answers' F1 scores of different methods. \textbf{Bottom:} the distribution of perplexity for the outputs.}
    \label{fig:intro}
\end{figure}

To further investigate this limitation, we analyze the framework classification proposed in FlashRAG~\cite{flashrag} and select four representative RAG methods, each corresponding to a distinct pipeline type: Branching, Iterative, Loop, and Agentic. As illustrated in Figure~\ref{fig:intro}, we evaluate these methods on four benchmark tasks and observe the following key findings:

(1) \textbf{Lack of generalizability in single RAG pipelines.} The upper part of Figure~\ref{fig:intro} presents the aggregated performance of each method across three datasets. Notably, Branching-based approaches (\eg, RePlug~\cite{replug}) underperform in multi-hop reasoning tasks but excel in multiple-choice settings. A similar phenomenon can be observed in the Iterative method, reinforcing that each pipeline exhibits task-dependent performance biases .

\begin{figure*}
    \centering
    \resizebox{1\textwidth}{!}{
    \includegraphics{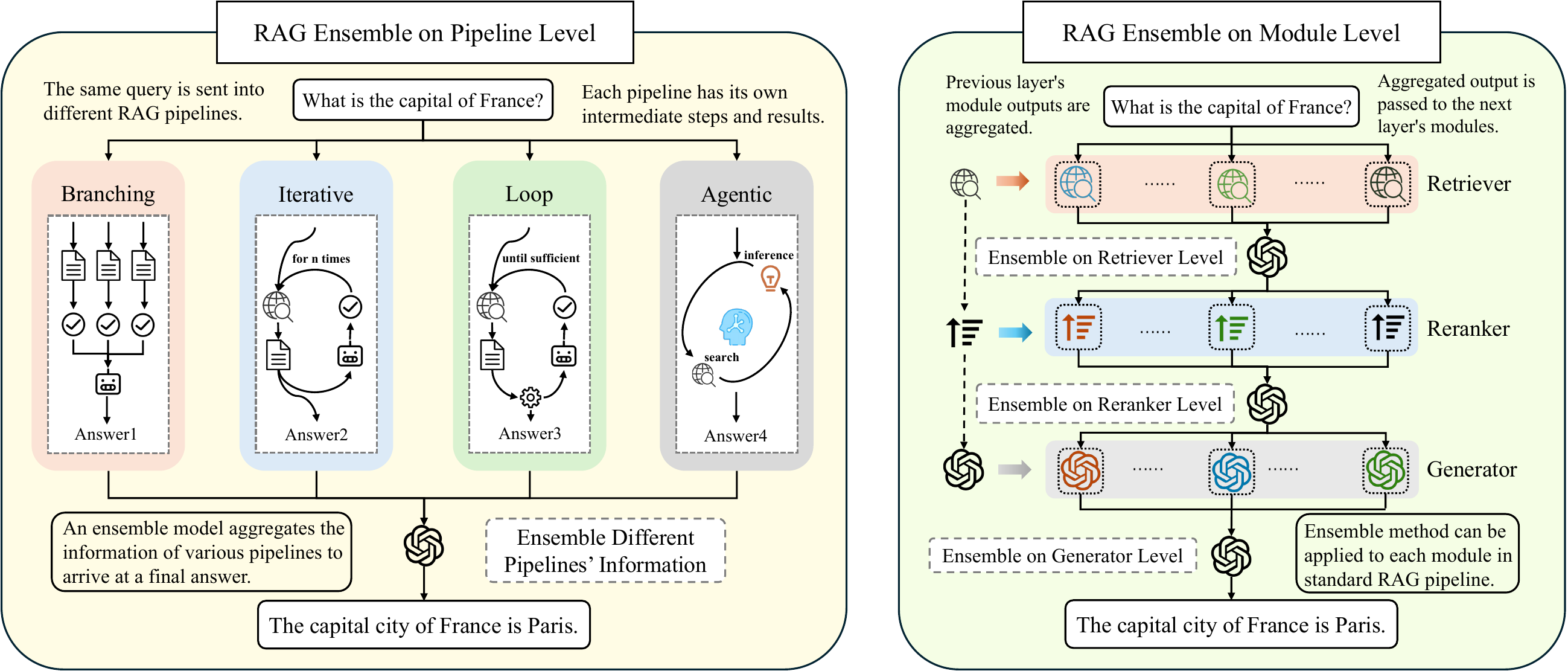}}
    \caption{The overall architecture of RAG Ensemble. We conduct a detailed study from the perspective of the overall pipeline level and the module level. The left framework represents ensemble from pipeline level, while the right framework represents ensemble from module level. }
    \label{fig:main_result}
\end{figure*}

(2) \textbf{Divergence in answer perplexity levels across pipelines.} By quantifying the perplexity of generated answers, we observe distinct density distributions for each method (lower part of Figure~\ref{fig:intro}). The more concentrated the distribution, the more stable the generated results. For instance, Loop-based methods yield lower answer perplexity (greater stability) on 2Wiki, but higher perplexity on WikiASP, reflecting task-specific confidence disparities.

These results collectively demonstrate that single-RAG systems struggle with task generalization, whether measured by performance or output perplexity. This motivates our core research question: How can we aggregate multiple RAG systems to enhance generalization capability for complex, heterogeneous tasks?

To address this issue, one intuitive approach is to perform adaptive fine-tuning on the model to enhance its ability in RAG tasks. However, such methods may interfere with the model's inherent capabilities and come with higher training costs. Another common strategy is to treat the model as a router, selecting the optimal single RAG system's answer and discarding the remaining systems. However, we consider that the unselected answers may still contain valuable information for the task. Recent research has begun to explore component ensemble methods. Some studies suggest that meta-search engines, by aggregating results from multiple search engines, can provide more relevant information~\cite{metasearch1,metasearch2}. Additionally, numerous studies focus on model-level ensemble strategies. We argue that, compared to routing methods, ensemble strategy can better make full use of the useful information in each subsystem, improving the quality of the final results. However, existing methods mainly focus on multi-component ensemble on single level, while RAG tasks involve more complex input flows and system structures. Unfortunately, both in terms of theoretical modeling and mechanism explanation, there is still a significant lack of systematic research on ensemble across multiple RAG systems, which significantly limits its development and application.

To address this challenge, in this paper, we conduct a comprehensive and systematic study of ensemble methods based on RAG systems. Specifically, we perform an in-depth analysis of the RAG system ensemble method from theoretical analysis and mechanism analysis:
\begin{enumerate}[leftmargin=1em]
\item \textbf{From a theoretical analysis perspective:} We model the RAG system ensemble method on a non-Euclidean manifold. Through detailed derivation, we clarify the effectiveness of RAG system ensemble from the perspective of information entropy increase. As we know, this is the first work to model a system ensemble task from the perspective of information entropy.

\item \textbf{From a mechanism analysis perspective:} To achieve a comprehensive exploration of RAG ensemble, we conduct in-depth investigations of seven different research questions from both the pipeline and module levels. At the system level, we carefully select four different RAG pipelines (Branching, Iterative, Loop, and Agentic) for ensemble research. Additionally, we conduct ensemble experiments on closed-source RAG frameworks to further explore the characteristics of RAG ensemble. At the module level, we conduct experimental research on the retriever, reranker, and generator of the standard RAG framework. We carefully select three retrievers and five generation models for the experiments, and delve into the characteristics of applying generative rerankers to ensemble tasks. Moreover, our experiments cover a wide range of task sets, including single-hop tasks, multi-hop tasks, multiple-choice tasks, summarization tasks, and tasks in vertical domains, all of which have detailed ensemble analysis.
\end{enumerate}
Our main findings include:
\begin{itemize}[leftmargin=1em]
\item RAG ensemble demonstrates clear advantages in both the framework type and the granularity of the ensemble. This reflects the good generalizability of the RAG ensemble method.

\item In a significant portion of ensemble tasks, the RAG ensemble method exhibits scaling-up characteristics, meaning that increasing the external information has a notable positive impact on the final ensemble result. However, this characteristic also depends on the model's strong resistance to information interference.

\item The ensemble model shows a preference for certain groups of input information, and this preference becomes more pronounced as task difficulty increases.
\end{itemize}
\section{Related Work}
\paragraph{\textbf{Retrieval-Augmented Generation}}

Retrieval-Augmented Generation (RAG) improves generation quality by integrating retrieved external knowledge~\citep{LewisPPPKGKLYR020,GuuLTPC20,liverag,abs-2308-07107}. Mainstream RAG methods follow a ``retrieve-then-read'' paradigm, where the retrieval module supplies external knowledge as context for generation models to produce output~\citep{replug,RamLDMSLS23,BorgeaudMHCRM0L22}. Recently, numerous improved RAG paradigms have emerged based on standard RAG technology~\cite{RAG3,flashrag,modularRAG2}. Foundation studies have focused on capturing more relevant and high-quality retrieval documents through advanced query rewriting~\citep{WangYW23,GaoMLC23} and fine-grained reranking techniques~\citep{YoranWRB24,00110PCML0024,RECOMP}. Other branch of efforts have attempted to introduce efficient fine-tuning strategies~\citep{spring,autoif,followrag,mao-etal-2024-rag,inters} to unlock the strong information-capturing capabilities of generators within RAG systems. Moreover, recent works aims to integrate self-correction and auto-evaluation methods into RAG domain~\citep{dongragcritic}.

Recently, with the rise in popularity of agentic search methods~\citep{Search_o1,hira}, a series of approaches have attempted to further enhance large models' autonomy and deep information retrieval capabilities using reinforcement learning techniques. For example, Search-R1 and Research use reinforcement learning to empower the model with the ability to complete deep search tasks~\citep{search-r1,chen2025research}. WebThinker and Kimi-Researcher attempt to leverage reinforcement learning to enhance the model's report writing ability~\citep{webthinker,kimik2}. Tool-Star and ARPO try to expand more tools to enhance the model's deep search capabilities~\citep{toolstar,arpo}. Furthermore, a series of studies have attempted to extend this progress to the multimodal domain~\citep{wu2025mmsearchr1incentivizinglmmssearch,wemath}.

However, existing approaches mainly enhance individual RAG systems, lacking exploration of collaborative multiple RAG systems. Our paper aims to investigate the effectiveness of model ensemble in RAG scenarios and introduces the \ours{} paradigm, offering practical insights for real-world applications.

\paragraph{\textbf{Model Ensemble in LLM}} Ensemble of LLMs has significantly outperformed individual models by leveraging the strengths of different systems. Existing ensemble approaches can be mainly categorized into three types: (1) A series of studies have fine-tuned external routers to select the most suitable LLM for specific inputs, enabling model selection before inference~\citep{abs-2309-15789,zooter}; (2) Another branch of efforts involves multiple models processing inputs incrementally and combining their outputs during decoding, showcasing strong collaborative potential~\citep{abs-2402-14845,HoangKJ24}; and (3) Some researchers focus on allowing each model to process inputs independently, then selecting the best response~\citep{abs-2305-05176,Jiang0L23}. To further improve the efficiency of LLMs ensemble, some works employ input compression~\citep{JiangWLYQ23,abs-2403-17411,LiuLXWQ23} and speculative decoding~\citep{abs-2405-19715,YuanLHYZ24} to accelerate model inference. Unfortunately, these studies have not systematically examined the application of ensemble techniques in the RAG domain, and integrating information at the model level alone is insufficient to bridge system gaps. Our study makes the first attempt to integrate all external knowledge and outputs from different RAG systems to maximize performance.

\section{Methodology and Theoretical Analysis}
In this section, we introduce the framework of RAG ensemble, and then we theoretically analyze why combining multiple RAG systems can be effective. 

\subsection{RAG Ensemble}


Given a set of RAG systems $\{\mathcal{S}_1,\mathcal{S}_2,\cdots,\mathcal{S}_n\}$, the ensemble framework aims to synthesize the inputs and outputs of these systems to generate a response $Y$. Specifically, each standard RAG system $\mathcal{S}_i$ includes a retriever $R_i$ and a generator $G_i$.\footnote{Many advanced RAG systems have designed several additional modules to improve the systems' performance. In theoretical analysis, we only consider the two fundamental components, without loss of generality, to make the formulation more clear. Henceforth, we will follow this principle to simplify the modules for better understanding.} Upon receiving a user input $X$, the retriever $R_i$ retrieves relevant external knowledge, denoted as $D_i=R_i(X)$. Then, the generator $G_i$ generates a response $Y_i=G_i(X, D_i)$. When multiple RAG systems exist, all inputs and outputs can be collected as follows: 
\begin{align}
    {S} &= \{{S}_1, {S}_2, \cdots, {S}_n\}, \\
    {S}_i &= \{Y_i, D_i\}.
\end{align}
Finally, an ensemble model is employed to generate the final response as follows:
\begin{align}
    Y = f_\phi(X, S),
\end{align}
where $f_\phi(\cdot)$ denotes the ensemble model parameterized by $\phi$.

The core idea of RAG ensemble lies in the ability of the model to aggregate the information from multiple RAG systems. This paper mainly focuses on the simplest method, which directly embeds the raw information from multiple RAG systems into a prompt and then inputs it to an LLM for ensemble. This process can be represented as follows:
\begin{align}
    Y = f_\phi (\text{Prompt}(X, {S})).
\end{align}

The prompt we use for basic RAG ensemble is as follows:

\begin{tcolorbox}[colframe=yellow!75!black, colback=yellow!5!white, title=\textbf{RAG Ensemble Prompt}]
Here is a question and some external data from \{num\} systems' information: \\
System 1: \{system 1's information\} \\
System 2: \{system 2's information\} \\
System 3: \{system 3's information\} \\
......

Question: \{question\}

Your task is to answer the question based on the given information. \
You should first output your reasoning process and then provide the final answer. \
The output format of reasoning process and final answer should be enclosed within <think> </think> and <answer> </answer> tags, \
respectively, and the final answer should be enclosed within \\boxed{{}} with latex format, \
\ie, ``<think> reasoning process here </think><answer>\boxed{a\ final\ answer\ here} </answer>''. \
Only output your reasoning process in <think></think> and your answer in <answer> boxed\{\} </answer>, \
and do not output any other words.
\end{tcolorbox}

In the following, we will provide a detailed theoretical analysis of the rationale behind RAG ensemble.

\begin{figure}[t]
    \centering
    \begin{subfigure}[b]{0.45\linewidth}
        \centering
        \includegraphics[width=\linewidth]{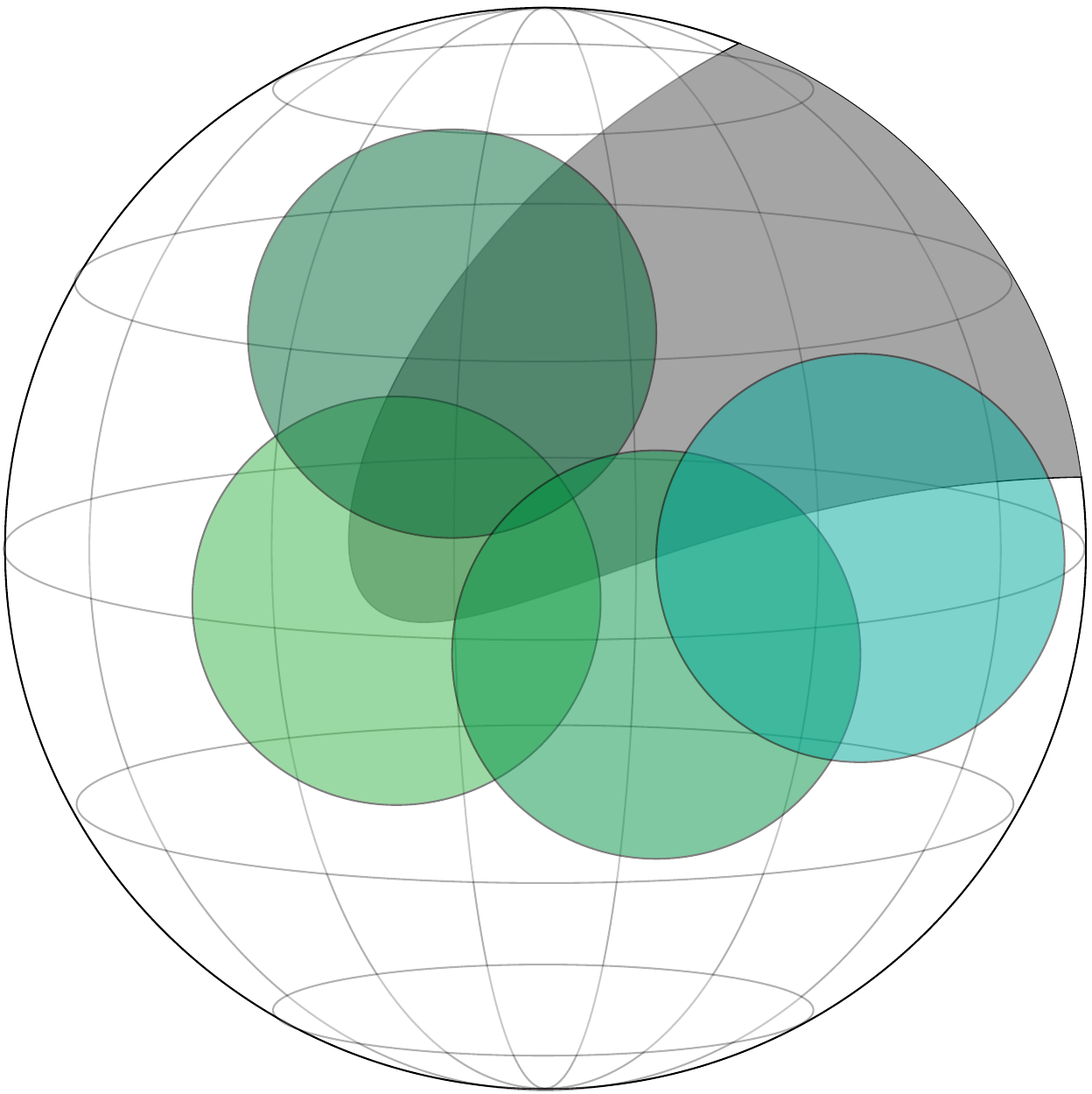}
        \caption{Before knowledge refinement}
        \label{fig:info_theory_2_1} 
    \end{subfigure}%
    \hfill
    \begin{subfigure}[b]{0.45\linewidth}
        \centering
        \includegraphics[width=\linewidth]{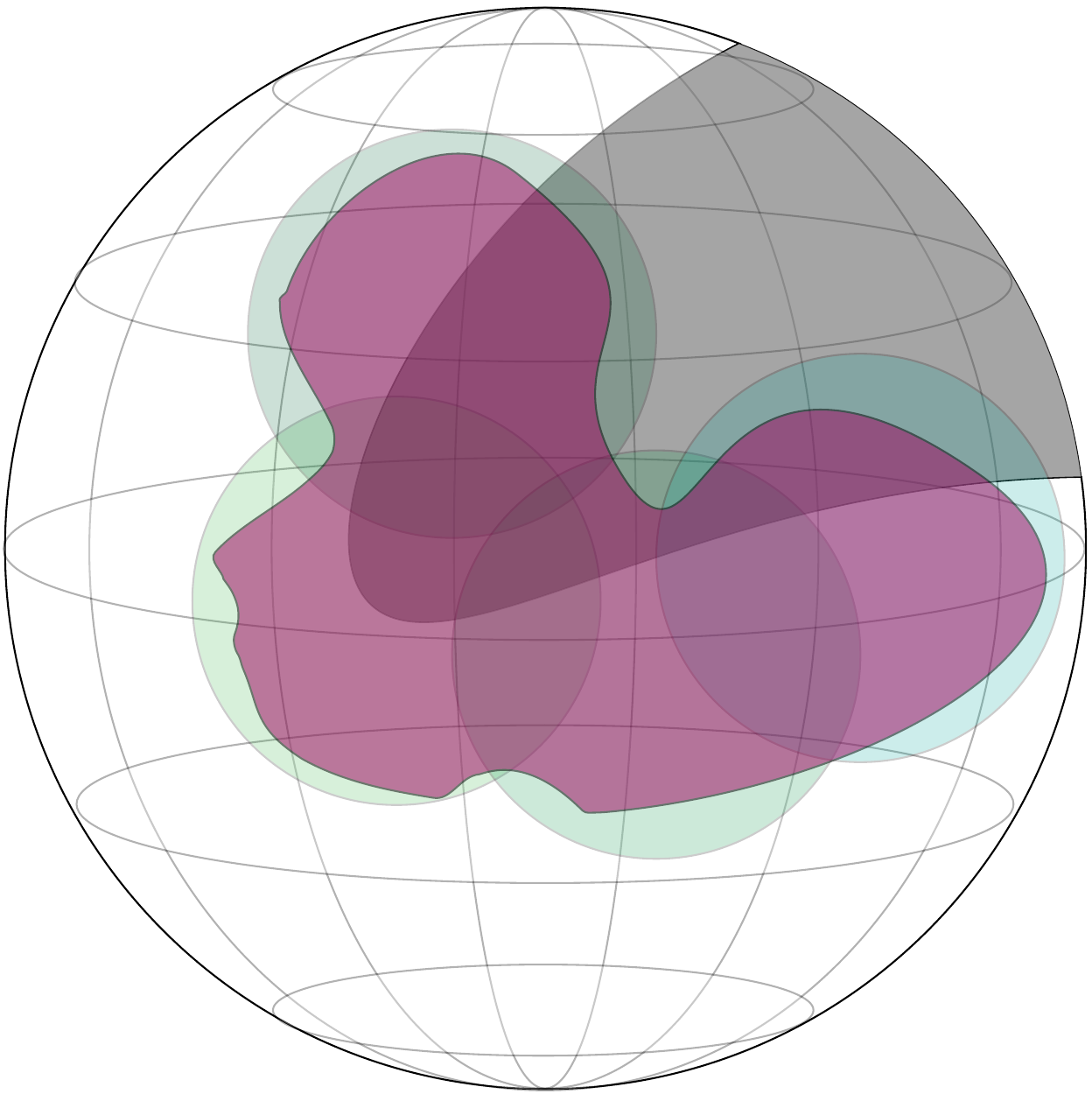}
        \caption{After knowledge refinement}
        \label{fig:info_theory_2_2} 
    \end{subfigure}
    \caption{Comparison of multi-RAG before and after knowledge refinement. Aggregating and analyzing the information from multi-RAG system (Figure~\ref{fig:info_theory_2_2}) is the key operation of ensemble.}
    \label{fig:info_theory_2}
\end{figure}

    


\subsection{Theoretical Analysis of RAG Ensemble}
\label{sec:thory}
In this section, we provide a detailed analysis of RAG ensemble from the perspective of theoretical modeling.
Given a probability distribution, information entropy describes the degree of uncertainty we have about an event. The higher the entropy, the more uncertain the system is, and the more information it contains. 
Mathematically, the entropy \( H(X) \) of a continuous random variable \( X \) with probability distribution \( P(x) \) can be shown as:
\begin{equation}
  \label{eq:informational entropy 6}
  H(X) = - \int p(x) \log p(x) \, dx.
\end{equation}

Inspired by the information bottleneck method~\cite{information-bottleneck-method,information-bottleneck-method2} in information theory, we consider that the key of the RAG ensemble framework lies in its ensemble of information from multiple individual RAG systems, reducing the information entropy of the final answer. As shown in Figure~\ref{fig:info_theory_2}, this process can be modeled on a non-Euclidean sphere.

As shown in Figure~\ref{fig:info_theory_2_1}, the entire white sphere represents the potential region for generating answers (\ie, the informational entropy of the answer), the gray area represents the useful information contained in the prompt, and each green circular area represents the external information introduced by an individual RAG system (including both useful and useless external knowledge). In Figure~\ref{fig:info_theory_2_2}, the purple area represents the useful information extracted by generator from all external knowledge. We consider that due to the introduction of useful knowledge from both the prompt and all individual RAG systems, the uncertainty in generating the answer is reduced. That is, RAG ensemble can aggregate useful information from multiple systems to generate a correct output. The useful knowledge of individual RAG systems and the useful knowledge after extraction from all systems can be obtained as follows:
\begin{align}
    e_i &= g_\phi \left[ q, (S_i) \right], \\
  \label{eq:informational entropy for blender knowledge extract}
    e^* &= g_\phi \left[ q, (S_1, S_2, \ldots, S_n) \right],
\end{align}
where \(e_i\) represents individual RAG's useful knowledge, \( e^* \) represents all the useful knowledge extracted, and \(g_\phi\) represents the process that the model \(\phi\) extracts useful information. Then, the process by which ensemble model reduces the information entropy of the generated answer can be expressed by the following formula:
\begin{align}
  H(Y|X, K) &= H(Y|X, e^*) \notag \\
  &= H(Y) -\underbrace{I(X, e^* ; Y)}_{\text{useful information}}, \notag \\
  I(X, e^*; Y) &= \underbrace{I(X, K; Y)}_{\text{all information}} - \underbrace{I(e_{\text{useless}})}_{\text{useless information}} \notag,
\end{align}
where \( K \) represents all the external knowledge provided, \( H(Y|X,K) \) represents the conditional entropy of the generated response when both the user prompt and external knowledge are introduced, \( H(Y) \) denotes the informational entropy of generating response. Moreover, \( H(Y|X, K) = H(Y|X, e^*) \) means that only useful information can help reduce the target response's entropy.  \( I(X, e^* ; Y) \), \( I(X, K; Y) \) represents the mutual information between the input and useful reference knowledge, as well as the mutual information between the input and all external knowledge, respectively. \( I(e_{\text{useless}}) \) represents the useless information in the external knowledge that is discarded after model analysis. Therefore, when external knowledge is introduced as a condition, the information entropy of the generated answer decreases due to the useful knowledge extracted by the ensemble model (\ie, the accuracy of generating answer improves).

For RAG ensemble system, if the external knowledge base does not contain conflicting information (\ie, the external knowledge base cannot simultaneously contain knowledge such as ``the sky is blue'' and ``the sky is green''), we can propose the following assumption:

\begin{tcolorbox}[colframe=blue!75!black, colback=blue!5!white, title=\textbf{Assumption}]
When performing ensemble tasks, an ideal model tends to refine the collected information in a direction that increases the amount of correct knowledge.
\end{tcolorbox}
\label{assumption for information increase}

Suppose we have an ideal ensemble model \(\phi^*\) that can perfectly extract useful information from the given external knowledge while ignoring all irrelevant information. In this case, the useful knowledge extracted from the input information of system \(i\) can be represented as follows:
\begin{equation}
  \label{eq:informational entropy 8}
  e_i = g_{\phi^*} \left( q, d_i, a_i\right).
\end{equation}

Meanwhile, the final useful information extracted from all subsystems can be expressed as follows:
\begin{equation}
  \label{eq:informational entropy 7}
  e^* = g_{\phi^*} \left( q, d_i, a_i, S_{{\textbackslash}i}  \right),
\end{equation}
here, \(S_{\backslash i}\) represents the input information excluding system \(i\).

When more system information is obtained, for the initial useful information \(e_i\), the ensemble model can handle it in the following two ways:

(1) \(\phi^*\) retains all the information \(e_i\). Due to the introduction of information from other systems, the total information received by the ensemble model is at least not worse than the information received solely by the \(i\)-th system. In other words, \(e_i\) is included in \(e^*\), which means:
\begin{equation}
\begin{aligned}
  \label{eq:informational entropy 9}
  I_1 &= I(q, e^*; a) \\
   &= I(q, e_i; a) + I(q, e_{{\textbackslash}i}^*; a),
\end{aligned}
\end{equation}

\(e_{{\textbackslash}i}^*\) represents the useful information from other systems excluding system \(i\). 

(2) If \(\phi^*\) only extracts partial information from \(e_i\), we define \(e_i^*\) represents the useful knowledge of system \(i\) after refinement, then the \(e_i^*\) can be represented as:
\begin{equation}
  \label{eq:informational entropy 10}
  I(q, e_i^*; a) = I(q, e_i; a) - I(q, e_{\text{i}}^{useless}; a),
\end{equation}
here, \(e_{\text{i}}^{useless}\) represents the part of the knowledge \(e_i\) that the model considers useless after final refinement. At this point, the information entropy of generating the answer can be expressed as:
\begin{equation}
\begin{aligned}
  \label{eq:informational entropy 11}
  I_2 &= I(q, e^*; a) \\
   &= I(q, e_i^*; a) + I(q, e_{{\textbackslash}i}^*; a)\\
  &=I(q, e_i; a) - I(q, e_{\text{i}}^{useless}; a)\\
  &+ I(q, e_{{\textbackslash}i}^*; a).
\end{aligned}
\end{equation}

At the same time, based on the assumption before, if the model chooses to discard some of the information from system \(i\), it must believe that other systems can provide more useful information, such that the total amount of effective information after refinement is greater than the information from system \(i\) alone, that is:
\begin{equation}
  \label{eq:informational entropy 12}
  I(q, e_{{\textbackslash}i}^*; a) \geq I(q, e_{\text{i}}^{useless}; a).
\end{equation}

Based on all the derivations above, we can conclude the following relationship:
\begin{equation}
\begin{aligned}
  \label{eq:informational entropy 13}
  H(a|q, e^*) &= H(a) - \min(I_1, I_2) \\
    &\leq H(a) - I(q, e_i; a) \\
    &= H(a|q, e_i),
\end{aligned}
\end{equation}


So the ensemble knowledge contains more useful information, and it helps reduce the information entropy in generating answers. 
Therefore, we consider that the process of ensemble can \textbf{introduce more helpful information than single system}. This is the core of RAG ensemble. In the following sections, we conduct a large number of experiments, clearly demonstrating the effectiveness of RAG ensemble.

\section{Experiments}
\label{experiments}

As shown in Figure~\ref{fig:main_result}, we conduct experiments to explore the RAG ensemble framework in depth. Based on the research in FlashRAG~\cite{flashrag}, we divide our experiments into ensemble at the pipeline level and at the module level. From the perspective of tasks, we carefully select datasets based on Wikipedia and MS MARCO~\cite{msmarco}. We explore the following research questions regarding ensemble at the pipeline and module levels:
\paragraph{\textbf{(1) Pipeline Level:}} 

\begin{itemize}[leftmargin=*]
\item \textbf{RQ1}: Does aggregating different pipelines effective? ~($\S$\ref{sec:results of ensemble on pipeline level})
\item \textbf{RQ2}: Is the ensemble method still effective when aggregating closed - source model pipelines? ~($\S$\ref{sec:results of ensemble on pipeline level})
\item \textbf{RQ3}: Is there a scaling - up phenomenon when aggregating? ~($\S$\ref{sec:scaling up})
\item \textbf{RQ4}: Does the ensemble model show a preference when aggregating? ~($\S$\ref{rag ensemble preference visualization})
\end{itemize}

\paragraph{\textbf{(2) Module Level:}} 

\begin{itemize}[leftmargin=*]
\item \textbf{RQ5}: When aggregating different generators' results, what is the performance of the ensemble framework? ~($\S$\ref{generator-level ensemble analysis})
\item \textbf{RQ6}: Can aggregating different retrievers' results help improve the performance? ~($\S$\ref{retriever level ensemble analysis})
\item \textbf{RQ7}: Is the ensemble method still effective when aggregating different rerankers? ~($\S$\ref{reranker level ensemble analysis})
\end{itemize}

\subsection{Datasets}
\label{datasets}
For the Wikipedia-based datasets, we carefully select four datasets: (1) TriviaQA~\cite{triviaqa}, a reading comprehension dataset containing a lot of triples, (2) 2WikiMultihopQA~\cite{2wiki}, a dataset containing multi-hop paths, (3) ARC~\cite{ARC}, a multiple-choice dataset containing real scientific questions, and (4) WikiASP~\cite{wikiasp}, a summarization generation dataset. They correspond to single-hop, multi-hop, multiple-choice, and open-domain summarization tasks, respectively. For each dataset, we randomly select 500 samples for evaluation. For the MS MARCO-based datasets, we follow the approach in RAG-Studio~\cite{rag_studio} and choose six vertical domains: Biomedical, Computing, Film, Finance, Law, and Music, with a maximum of 1000 test samples selected from each dataset. For the experiments on pipeline level, we use Wikipedia-based datasets, while for the module level, we use MS MARCO-based datasets. In the subsequent sections, we use ``2Wiki'' as the abbreviation for 2WikiMultiHopQA dataset.

\subsection{Baselines}
\label{sec:baselines}

We select representative methods from each of four RAG frameworks: \textbf{Branching, Iterative, Loop} and \textbf{Agentic} methods. For the Agentic RAG technology, we conduct experiments with both prompting-based models and reinforcement learning models, respectively. For Branching framework, we choose Replug method~\citep{replug}; for the Iterative framework, we choose Iter-RetGen method~\citep{modularRAG2}; for the prompting-based Agentic framework, we choose Search-o1 method~\citep{Search_o1}; and for the RL-based Agentic framework, we choose R1-Searcher method~\citep{r1searcher}. Due to open-source models' type limitations, for the Loop framework, we use Self-RAG with Llama3-8B-Instruct and FLARE with Qwen2.5-7B-Instruct as backbone models~\citep{SELF-RAG,FLARE}.

In order to fully explore the characteristics of the ensemble method, in addition to RAG ensemble generation—method that directly generates an answer, we also set up a RAG ensemble selection method. Specifically, we use the same model as the one used for generation, but let it select the optimal answer from the given candidate answers. We call this method RAG ensemble selection.

\subsection{Experimental Settings}
\label{sec:experimental settings}
In pipeline level experiments, we choose the Llama3-8B-Instruct\cite{llama3} for the main experiment. In addition, we also select Llama3.1-8B-Instruct~\cite{llama3}, Qwen2.5-7B-Instruct~\cite{qwen2.5}, Qwen2-7B-Instruct~\cite{qwen2-7b}, and Mistral-7B-Instruct-v0.3~\cite{mistral} for experiments related to base model ablation study. Due to the limitations of open-source models, for the Self-RAG method, we only use Llama3-8B-Instruct as base model. For retrieval step, we use e5-base~\cite{E5} model as the retriever, and retrieve top-5 documents from Wikipedia dumps\footnote{https://archive.org/details/enwiki-20181220}. We use exact match (EM) and F1 score for 2WikiMultiHopQA, TriviaQA, and ARC datasets as metrics, and use F1 score and ROUGE-L score for WikiASP dataset.

In the module level experiments, we use Llama3-8B-Instruct, Llama3.1-8B-Instruct, Qwen2-7B-Instruct, and Mistral-7B-Instruct-v0.3 for generator level and reranker level ensemble. For the experiment on retriever level ensemble, we choose e5-base model, contriever\cite{contriever}, and BM25\cite{bm25} as our experiment's retrievers. k1 parameter in BM25 algorithm is set to 1.5, and b parameter is set to 0.75.

\begin{table*}[t]
\small
\centering
\caption{Overall results on 4 QA datasets. The top two results in each backbone model's baseline group are highlighted in bold and underlined. ``Llama'' means the backbone model is Llama3-8B-Instruct, and ``Qwen'' means the backbone model is Qwen2.5-7B-Instruct. The average score in the last column we report is the mean of F1 score across the four datasets. 2Wiki. (2WikiMultiHopQA).}

\setlength{\tabcolsep}{4mm} 
\renewcommand{\arraystretch}{1.2} 
\begin{tabular}{lcccccccccc}
\toprule
\multirow{2}[2]{*}{\textbf{Method}} & \multirow{2}{*}{\textbf{Pipeline Type}} & \multicolumn{2}{c}{\textbf{2Wiki}} & \multicolumn{2}{c}{\textbf{TriviaQA}} & \multicolumn{2}{c}{\textbf{ARC}} & \multicolumn{2}{c}{\textbf{WikiASP}} & \multirow{2}[2]{*}{\textbf{Avg.}} \\
\cmidrule(lr){3-4} \cmidrule(lr){5-6} \cmidrule(lr){7-8} \cmidrule(lr){9-10}
 & & \textbf{EM} & \textbf{F1} & \textbf{EM} & \textbf{F1} & \textbf{EM} & \textbf{F1} & \textbf{F1} & \textbf{Rouge-L} & \\
\midrule

\multicolumn{10}{l}{\textit{\textbf{Different RAG Baselines with Llama}}} \\
RePlug & Branching & 13.6 & 20.0 & 53.4 & 61.3 & \textbf{84.4} & 62.4 & \underline{21.7} & 11.6 & 41.4 \\
Iter-RetGen & Iterative & 10.2 & 20.2 & \underline{58.0} & \underline{67.1} & 76.2 & 53.0 & \textbf{23.1} & 12.3 & 40.9 \\
Self-RAG &Loop & 12.2 & 24.0 & 39.6 & 53.3 & 67.0 & 57.0 & 15.7 & 9.8 & 37.5 \\
Search-o1 & Agentic (Prompt) & 15.2 & 23.6 & 50.6 & 59.3 & 26.8 & 34.4 & 16.0 & 5.8 & 33.3 \\
R1-Searcher & Agentic (RL) & \underline{51.0} & \underline{55.8} & \underline{58.0} & 65.8 & 64.4 & \textbf{67.6} & 17.7 & \underline{13.9} & \underline{51.7}\\
\midrule

\multicolumn{10}{l}{\textit{\textbf{Multiple Systems Integrated Methods with Llama}}} \\
\multicolumn{2}{l}{RAG Ensemble(Generation)} & \textbf{52.5} & \textbf{56.0} & \textbf{63.6} & \textbf{72.2} & \underline{82.6} & \underline{62.5} & 18.7 & 10.1 & \textbf{52.4} \\
\multicolumn{2}{l}{RAG Ensemble(Selection)}  & 23.4 & 31.5 & 56.4 & 64.0 & 62.8 & 52.1 & 18.0 & \textbf{14.1} & 41.4 \\
\midrule

\multicolumn{10}{l}{\textit{\textbf{Different RAG Baselines with Qwen}}} \\
RePlug &Branching& 23.8 & 27.5 & 47.4 & 54.5 & 91.0 & 68.2 & 18.9 & 11.3 & 42.3 \\
Iter-RetGen &Iterative& 30.2 & 36.2 & \textbf{59.2} & \textbf{66.8} & 90.4 & 68.0 & \textbf{23.1} & 12.0 & 48.5 \\
FLARE & Loop & 20.2 & 25.1 & 41.0 & 47.6 & 79.2 & 58.8 & 17.4 & \underline{13.3} & 37.2\\
Search-o1 &Agentic (Prompt)& 32.0 & 39.0 & 55.0 & 63.2 & \textbf{93.0} & \underline{70.0} & 18.4 & 10.2 & 47.7 \\
R1-Searcher & Agentic (RL) & \underline{51.8} & \underline{57.9} & 54.8 & 64.6 & 88.2 & 68.4 & 6.1 & 4.7 & \underline{49.3}\\
\midrule

\multicolumn{10}{l}{\textit{\textbf{Multiple Systems Integrated Methods with Qwen}}} \\
\multicolumn{2}{l}{RAG Ensemble(Generation)}& \textbf{54.8} & \textbf{64.5} & \underline{55.2} & \underline{65.8} & \underline{91.8} & \textbf{70.4} & \underline{19.8} & \textbf{14.3} & \textbf{55.1}\\
\multicolumn{2}{l}{RAG Ensemble(Selection)} & 37.2 & 43.5 & 54.5 & 63.2 & 90.4 & 69.0 & 17.1 & 12.6 &  48.2\\

\bottomrule
\end{tabular}
\label{tab:main_result}
\end{table*}

\subsection{Main Results on RAG Pipeline Ensemble}

In this section, we perform a comparative performance analysis of RAG methods from both \textbf{open-source} and \textbf{closed-source} perspectives.
\label{sec:results of ensemble on pipeline level}

\begin{figure}[t]
    \centering
    \includegraphics[width=\linewidth]{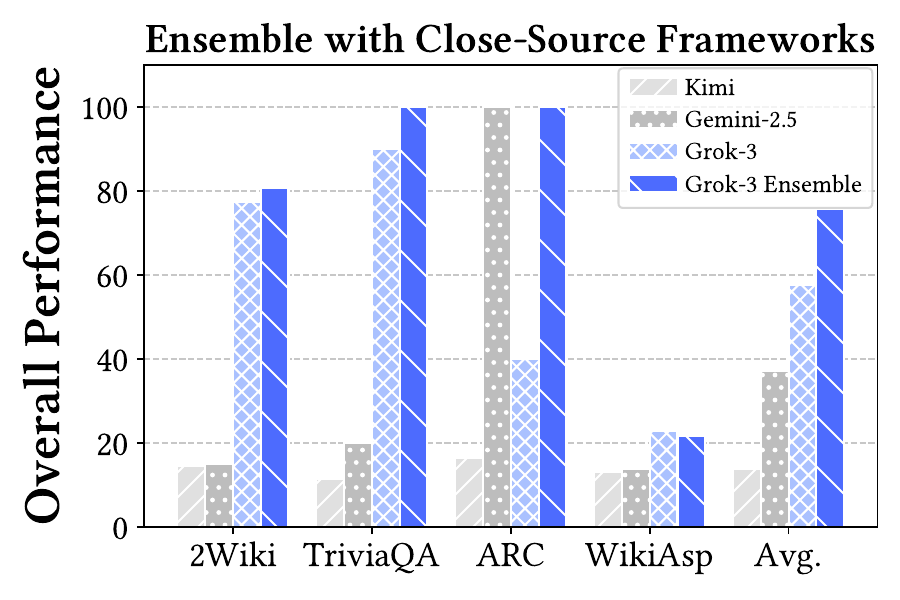}
    \caption{The performance of RAG Ensemble with different close-source frameworks.}
    \label{fig:close}
\end{figure}

\subsubsection{Analysis of Open-Source Methods.} Table~\ref{tab:main_result} presents the main results of our proposed RAG ensemble method. Overall, the ensemble method consistently demonstrates superior average performance across different backbone models in average performance. A more detailed analysis yields the following key observations:

\begin{itemize}[leftmargin=1em]
\item \textbf{Different methods excel in different tasks.} The effectiveness of different methods shows considerable variation depending on tasks. For instance, when using LLaMA as the backbone, RePlug achieves the best EM score on the ARC multiple-choice task, while Iter-RetGen outperforms RePlug and other variants on F1 score in the WikiASP summarization task. Similarly, R1-Searcher performs worse than RePlug on ARC and WikiASP but performs better on 2Wiki and TriviaQA. These results support our claim that no single RAG framework is universally optimal—different methods are more suited to specific task types.
\item \textbf{The ensemble method's stability and performance are superior to those of the baseline method on average.} Regardless of the backbone model, the performance of the ensemble method can reach the optimal or sub-optimal level on most metrics, and it achieves the best result in the average F1 score. This highlights the effectiveness of the ensemble method. In addition, whether integrating pure prompting-based RAG methods (such as Iter-RetGen and FLARE) or integrating RAG methods with training (such as Self-RAG and R1-Searcher), the ensemble method can also achieve good results. This highlights the stability of the ensemble method.
\item \textbf{Generative ensemble outperforms selective ensemble in RAG.} Compared to selecting a single best candidate answer, fusing multiple answers through generative ensemble methods leads to superior performance. For instance, under the LLaMA-base setting, selection-based approaches underperform even some individual baselines, falling 27.6 points behind R1-Searcher on 2Wiki's EM score and 21.6 points behind RePlug on ARC's EM score. This performance gap may be attributed to the limited quality of selected candidates and inadequate comparative reasoning by the model. This result highlights the limitation of relying solely on a single RAG framework.
\end{itemize}

\begin{figure}[t]
    \centering
    \includegraphics[width=\linewidth]{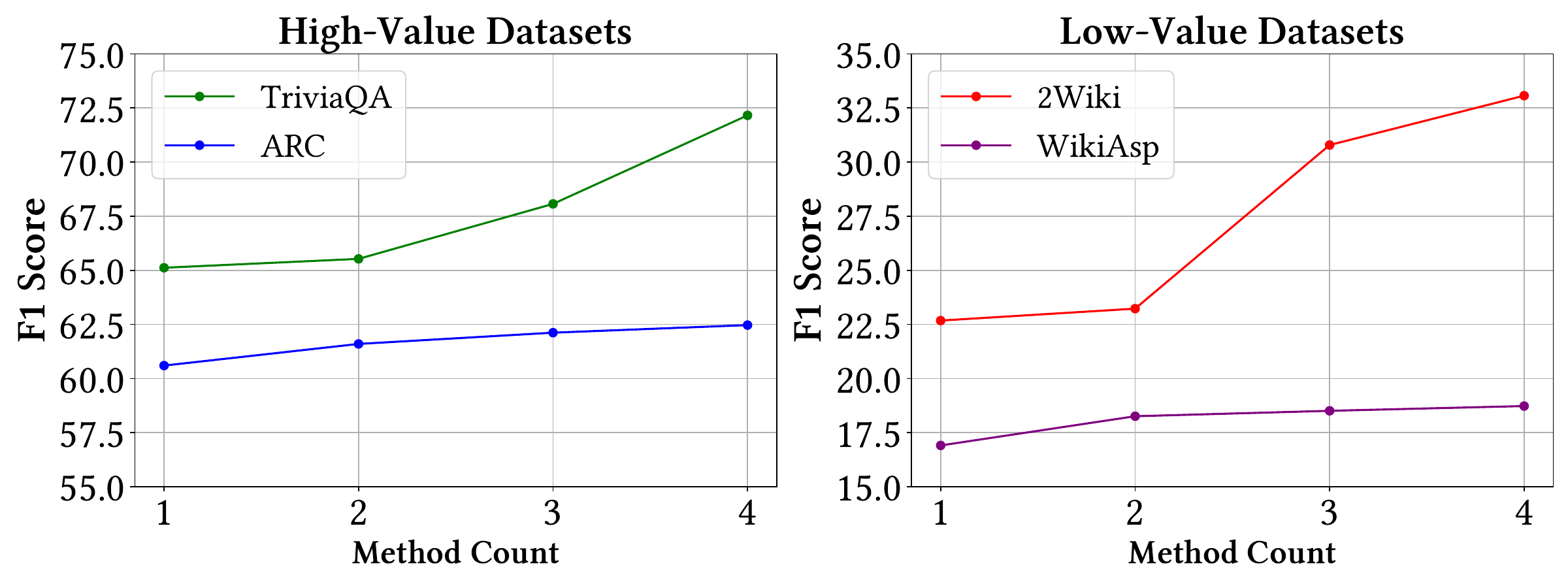}
    \caption{Ensemble of different scales at pipeline level.}
    \label{fig:scaling laws qwen}
\end{figure}

\subsubsection{Analysis of Close-Source Methods.} To comprehensively assess the applicability of the RAG Ensemble framework, we further conduct experiments on closed-source models. In this experiment, we select three closed-source models—Kimi, Gemini-2.5\cite{gemini}, and Grok-3—for RAG inference, using Kimi as the ensemble model. Then, we randomly sample 20 subsets from each of four datasets for preliminary evaluation.

As shown in  Figure~\ref{fig:close}, results demonstrate that the RAG Ensemble framework remains effective when applied to closed-source models. While individual models may achieve strong performance on specific tasks, they often underperform on others. This observation supports our claim in the introduction: the performance of a single RAG framework may vary significantly across tasks due to multiple factors. By aggregating outputs from different models, the RAG Ensemble effectively mitigates inter-system performance variance, thereby overcoming the limitations of any single model.

\subsection{Ensemble System-Scale Scalability Analysis}
\label{sec:scaling up}
Building on the theoretical analysis in section~\ref{sec:thory}, the effectiveness of the RAG Ensemble framework primarily stems from the complementary information provided by different RAG systems, which helps mitigate the limitations of individual systems. To validate the hypothesis that incorporating more systems contributes to performance improvement, we investigate the scalability of the RAG system ensemble. In detail, we vary the number of aggregated systems from 1 to 4. For each setting, we enumerate all possible system combinations and report the average performance across combinations as the overall result for that scale. 

As shown in Figure~\ref{fig:scaling laws qwen}. Results indicate a clear upward trend in performance as the number of aggregated systems increases, revealing a strong positive correlation between ensemble size and reasoning effectiveness. This further supports our hypothesis: integrating information from more systems enhances final inference performance, demonstrating the framework's strong scalability.

\begin{figure}[t]
    \centering
    \includegraphics[width=\linewidth]{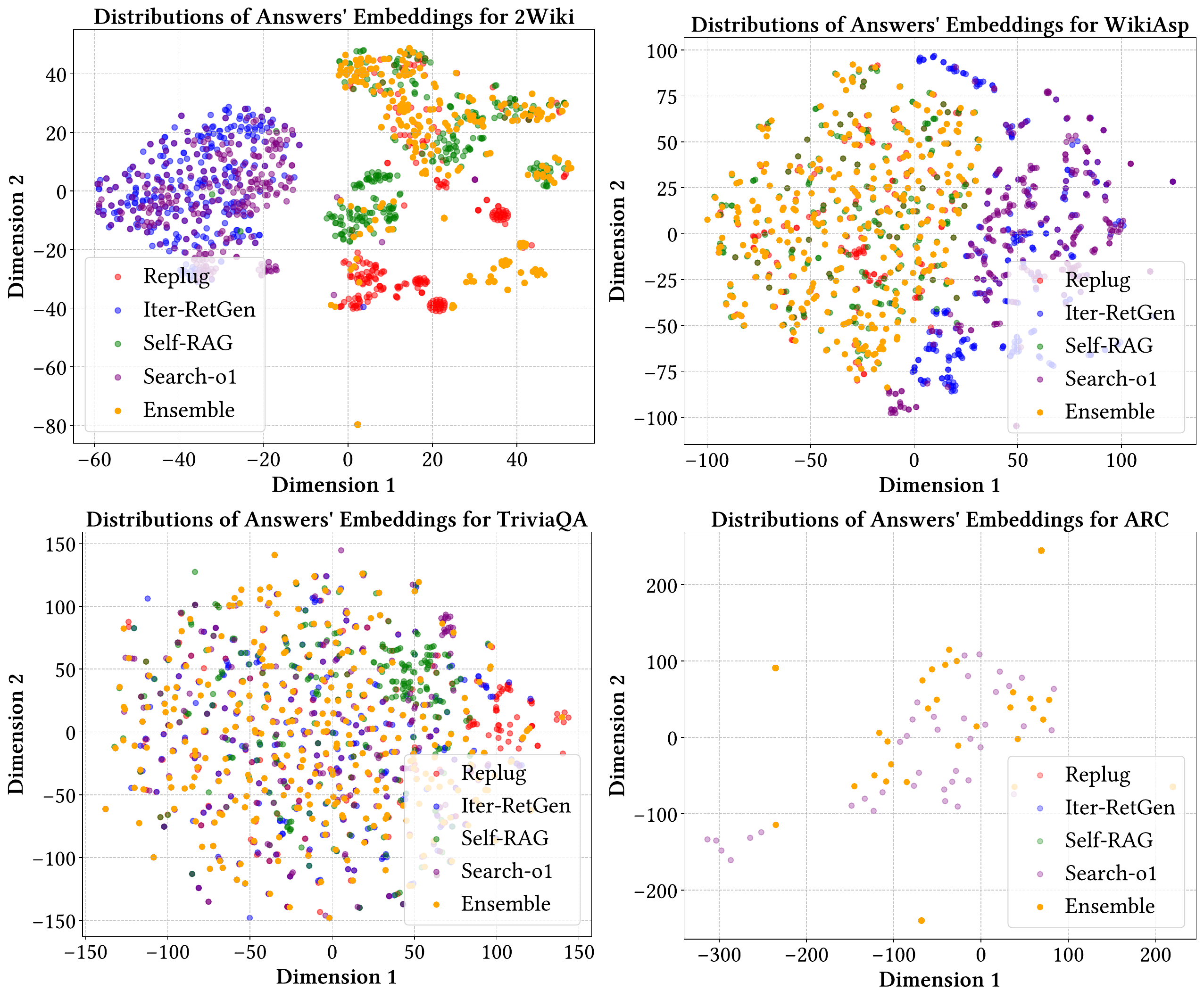}
    \caption{Distributions of ensemble answers and sub-answers.}
    \label{fig:t_sne}
\end{figure}




\subsection{RAG Ensemble Preference Visualization}
\label{rag ensemble preference visualization}
In preliminary experiments, we find that the ensemble method's result is slightly lower than the best-performing baseline on certain datasets. We hypothesize that this may result from the ensemble model exhibiting preferences toward specific subsystems during answer generation. To validate this assumption, we further analyze the distributional differences between subsystem outputs and ensemble responses. Specifically, we use the BGE model~\cite{bge} to encode answers generated by the ensemble and each subsystem, and apply t-SNE to visualize the embeddings. As shown in Figure~\ref{fig:t_sne}, the visualization reveals two key insights:

\begin{itemize}[leftmargin=1em]
\item \textbf{The ensemble model exhibits clear preference.} On the 2Wiki and WikiASP datasets, the embedding distribution of ensemble answers is more closely aligned with those of Self-RAG and RePlug, while differing significantly from Iter-RetGen and Search-o1. This indicates that, for these tasks, the ensemble model tends to rely more on information provided by Self-RAG and RePlug.
\item \textbf{The degree of ensemble preference correlates with subsystem performance.} On the TriviaQA and ARC datasets, where all subsystems achieve relatively high and comparable performance, the ensemble model shows no strong bias toward any particular method. In contrast, for more challenging tasks such as 2Wiki and WikiASP, where subsystem performance varies widely, the ensemble model tends to favor the stronger-performing methods. This suggests that ensemble preference is influenced, to some extent, by the performance distribution of the subsystems.
\end{itemize}

\begin{table*}[t]
\small
\centering
\caption{Overall results for generator level's ensemble on MS MARCO. The models we use here are all instruct versions. In the ensemble experiments, ``Llama'' means the ensemble model is \texttt{Llama3-8B-Instruct}, and ``Qwen'' means the ensemble model is \texttt{Qwen2.5-7B-Instruct}.}

\setlength{\tabcolsep}{1mm} 
\begin{tabular}{lcccccccccccccc}
\toprule
\multirow{2}[2]{*}{\textbf{Backbone}} & \multicolumn{2}{c}{\textbf{Biomedical}} & \multicolumn{2}{c}{\textbf{Computing}} & \multicolumn{2}{c}{\textbf{Film}} & \multicolumn{2}{c}{\textbf{Finance}} & \multicolumn{2}{c}{\textbf{Law}} & \multicolumn{2}{c}{\textbf{Music}} & \multicolumn{2}{c}{\textbf{Avg.}}\\
\cmidrule(lr){2-3} \cmidrule(lr){4-5} \cmidrule(lr){6-7} \cmidrule(lr){8-9} \cmidrule(lr){10-11} \cmidrule(lr){12-13} \cmidrule(lr){14-15}
 & \textbf{F1} & \textbf{Rouge-L} & \textbf{F1} & \textbf{Rouge-L} & \textbf{F1} & \textbf{Rouge-L} & \textbf{F1} & \textbf{Rouge-L} & \textbf{F1} & \textbf{Rouge-L} & \textbf{F1} & \textbf{Rouge-L} & \textbf{F1} & \textbf{Rouge-L}\\
\midrule

\multicolumn{10}{l}{\textit{\textbf{Different Base Models for Standard RAG}}} \\
Llama3-8B & 29.6 & 21.2 & \textbf{32.7} & 23.1 & \underline{45.9} & 42.2 & 28.9 & 23.2 & 31.1 & 25.6 & 45.4 & 40.0 & 35.6 & 29.2\\
Qwen2-7B & 29.5 & 21.2 & 30.1 & 24.0 & 38.8 & 33.7 & 29.2 & 22.7 & 31.7 & 21.8 & 40.6 & 34.5 & 33.3 & 26.3\\
Mistral-7B & 26.5 & 19.6 & 31.2 & \underline{24.9} & 38.4 & 34.6 & 26.5 & 20.6 & 30.1 & 24.5 & 36.7 & 31.5 & 31.6 & 26.0\\
Llama3.1-8B & 27.5 & 21.2 & 28.9 & 23.1 & 44.8 & 41.9 & 28.3 & 23.8 & 29.1 & 24.1 & 43.3 & 37.7 & 33.7 & 28.6\\
Qwen2.5-7B & 28.4 & 21.5 & 29.8 & 23.2 & 36.5 & 32.0 & 28.1 & 21.8 & 31.2 & 25.6 & 41.2 & 35.2 & 32.5 & 26.6\\
\midrule

\multicolumn{10}{l}{\textit{\textbf{Multiple Generators Ensemble}}} \\
Ensemble with Llama & \textbf{31.6} & \textbf{22.6} & 31.3 & 24.6 & \textbf{52.3} & \textbf{47.9} & \textbf{34.8} & \textbf{28.1} & \textbf{33.0} & \textbf{26.7} & \textbf{51.9} & \textbf{46.1}& \textbf{39.2} & \textbf{32.7} \\
Ensemble with Qwen & \underline{31.0} & \underline{22.3} & \underline{32.6} & \textbf{25.6} & \textbf{52.3} & \underline{47.5} & \underline{32.9} & \underline{26.0} & \underline{32.1} & \underline{26.1} & \underline{48.0} & \underline{42.4} &\underline{38.2} & \underline{31.7}\\

\bottomrule
\end{tabular}

\label{tab:msmarco_main_result}
\end{table*}

\begin{figure}[t]
    \centering
    \includegraphics[width=\linewidth]{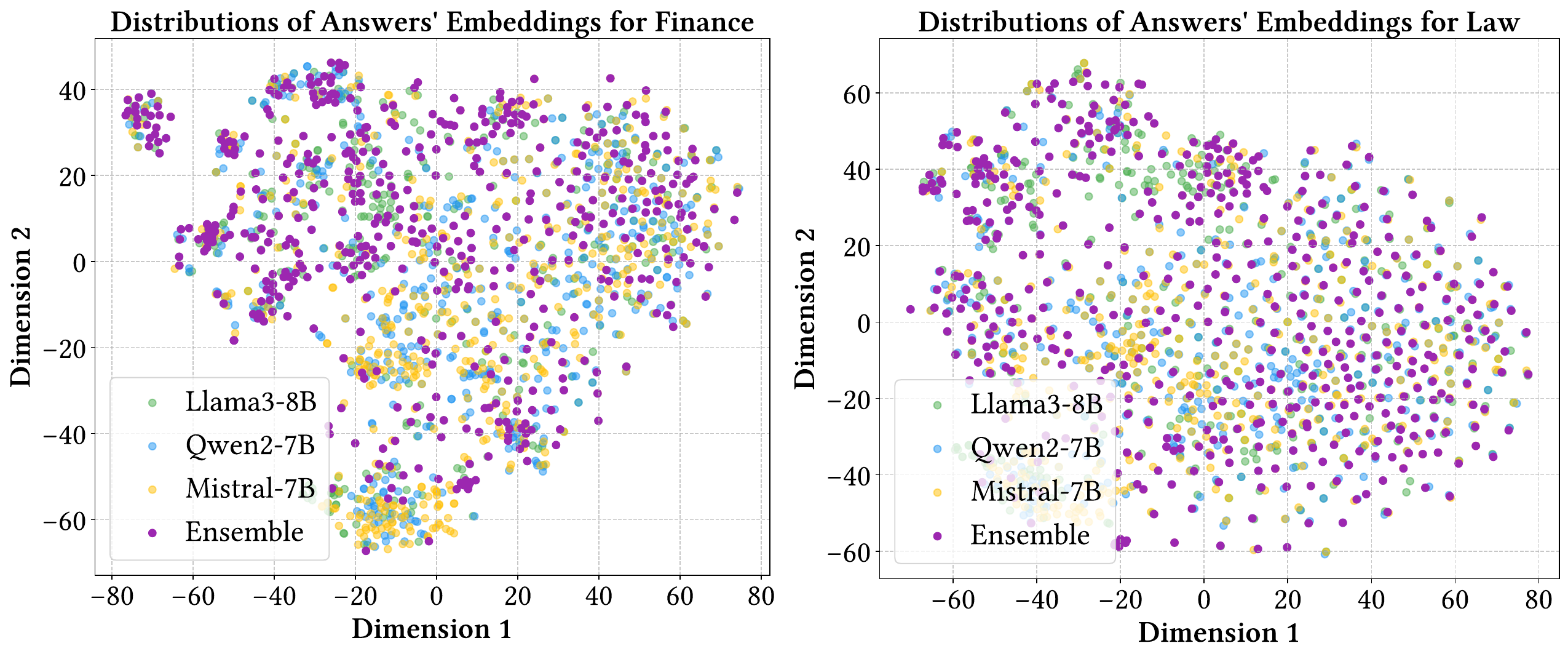}
    \caption{Distributions of ensemble answers and sub-answers on MS MARCO datasets.}
    \label{fig:t_sne_msmarco}
\end{figure}

\subsection{Generator-level Ensemble Analysis}
\label{generator-level ensemble analysis}

Starting from this section, we will focus on exploring the experiments on RAG ensemble at the module level.

\textbf{Overall Results.} To further evaluate the effectiveness of generator level's ensemble in RAG systems, we design an experiment specifically targeting generator ensemble. For each question, the reference documents remain fixed, while only the answer-generating model is varied. All generated outputs are then fed into an ensemble model to produce a final answer. In this setup, Llama3-8B-Instruct, Qwen2-7B-Instruct, and Mistral-7B-Instruct-v0.3 serve as the generators to be aggregated, while Llama3.1-8B-Instruct and Qwen2.5-7B-Instruct are used as ensemble models. As shown in Table~\ref{tab:msmarco_main_result}, our key findings are as follows:

\begin{itemize}[leftmargin=1em]
\item \textbf{Even under a fixed pipeline, aggregating outputs from different generators consistently yields strong performance gains.} In terms of the F1 score, ensemble using Llama3.1-8B-Instruct improves performance by 3.6\% over the best-performing single-generator baseline (Llama3-8B-Instruct), while Qwen2.5-7B-Instruct achieves a 2.6\% improvement. Across six vertical-domain tasks, the ensemble approach consistently reaches either the best or second-best results. This demonstrates ensemble framework's robustness across domains.

\item \textbf{Diversity among candidate answers plays a critical role in enhancing ensemble performance.} We compare the ensemble framework with a standard RAG setup where the same ensemble model directly generates answers using the same reference documents. Despite identical inputs, the ensemble framework, enriched with subsystem outputs, achieves superior results. For instance, using Llama3.1-8B-Instruct, the aggregated output yields a 6.7\% improvement in average F1 score and a 4.1\% gain in ROUGE-L compared to its standard RAG counterpart. This suggests that different generators offer complementary perspectives on the same evidence, which the ensemble model effectively synthesizes into more accurate answers. These results highlight the importance of answer diversity in improving final output quality.
\end{itemize}

\begin{figure}[t]
    \centering
    \includegraphics[width=\linewidth]{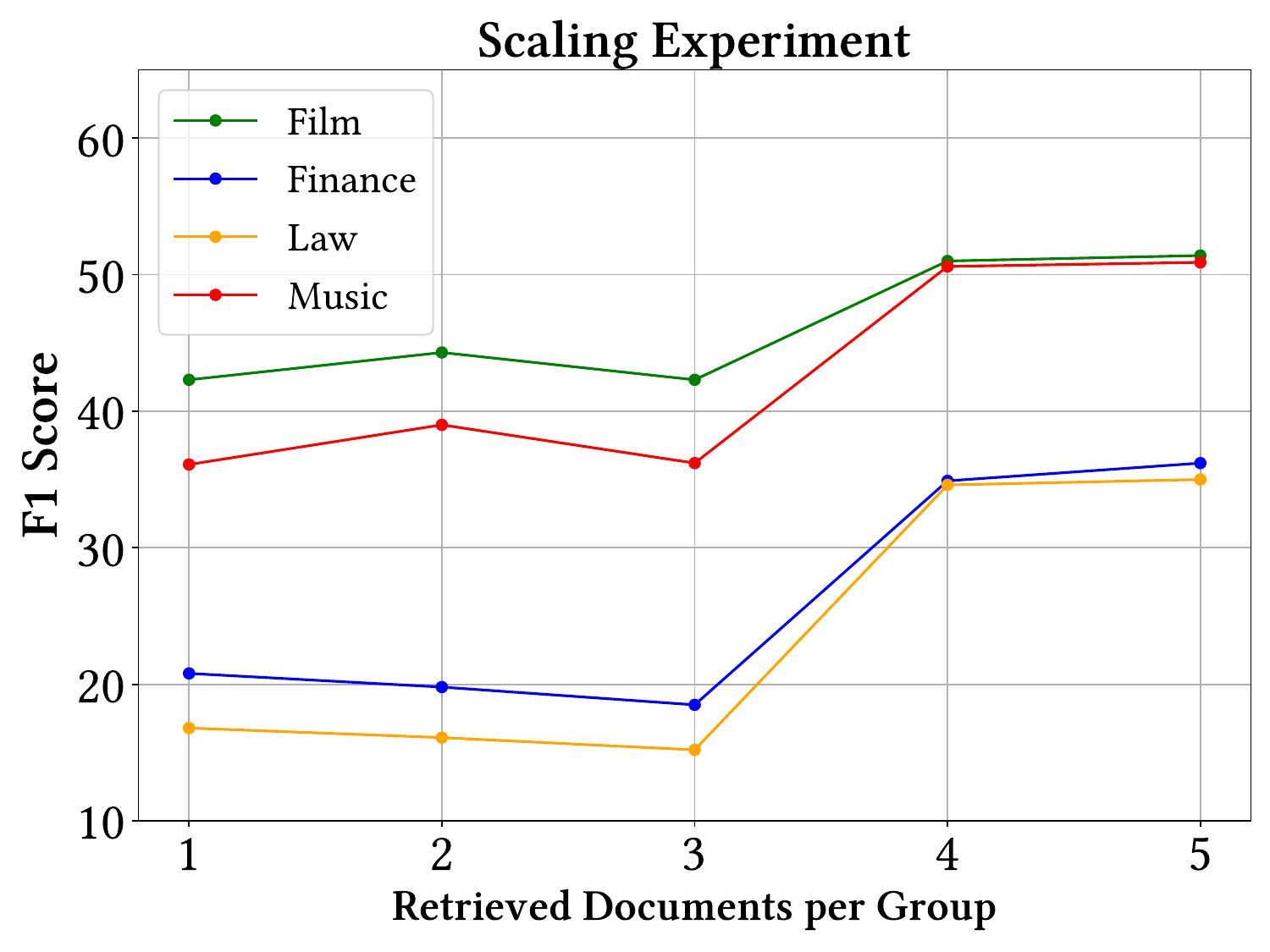}
    \caption{F1 Score for retriever level's ensemble on MS MARCO.}
    \label{fig:retriever}
\end{figure}

\textbf{Preference Visualization Analysis.} Furthermore, we analyze the preference of the ensemble model in domain-specific tasks. We select the Finance and Law datasets for this analysis, with the results shown in the Figure~\ref{fig:t_sne_msmarco}.
The results indicate that the ensemble model exhibits no clear preference in these two domains, as the embeddings of answers from individual systems and the aggregated outputs are relatively evenly distributed. This aligns with our earlier hypothesis that significant preference only emerges when the performance gap among subsystems is large. In the Finance and Law tasks, the performance of different generators is relatively similar, leading to low distinguishability among their outputs and, consequently, no strong preference in the ensemble model.

\subsection{Retriever-level Ensemble Analysis}
\label{retriever level ensemble analysis}
To fully explore the generalization of ensemble method in the retriever module, we conduct experiments using different retrievers on the MS MARCO datasets. We use three retrievers: BM25, Contriever, and E5, with each retriever recalling the top-5 documents. Additionally, we also perform scaling-up experiments at the document quantity level.

\begin{table}[t]
\small
\centering

\caption{Different base models' results and ensemble results under the same pipeline on TriviaQA dataset. All the base models are instruct versions.}
\setlength{\tabcolsep}{1mm} 
\begin{tabular}{lcccccc}
\toprule
\textbf{Method} & \textbf{Biomedical} & \textbf{Computing} & \textbf{Film} & \textbf{Finance} & \textbf{Law} & \textbf{Music} \\

\midrule
BM25 & 17.3 & 16.0 & 38.0 & 18.1 & 15.2 & 34.9\\
Contriever & 16.8 & 16.1 & 39.0 & 19.0 & 15.9 & \underline{37.5}\\
E5 & \underline{18.2} & \underline{18.3} & \underline{42.4} & \underline{19.5} & \underline{17.8} & 37.3\\
Ensemble & \textbf{33.1} & \textbf{34.4} & \textbf{51.4} & \textbf{36.2} & \textbf{35.0} & \textbf{50.9}\\
\bottomrule
\end{tabular}
\label{tab:scaling up retriever}
\end{table}

The experimental results are shown in Table~\ref{tab:scaling up retriever}. For all datasets, the ensemble method at the retriever level is still superior compared to the single-retriever RAG method. This highlights the strong generalization ability of the RAG ensemble at the retriever level. Subsequently, we continue with the scaling-up experiment on the number of retrieved documents, and the experimental results are shown in Figure~\ref{fig:retriever}.

In the early stages of documents increase, the improvement in ensemble performance is relatively small (Film, Music), and in some cases, the performance even decreases (Finance, Law). However, as the documents' number continue to increase, a significant improvement in ensemble performance can be observed across all datasets. We consider that this is because the robustness of the model can have a significant impact on the effectiveness of ensemble task. In the early stages of documents increase, the ensemble model may not have been sufficiently robust due to the increase in redundant information. However, when the number of documents reaches a certain threshold, the valuable information might be more easily captured by the model due to its multi-perspective presentation, thus suppressing the noise and leading to a substantial performance improvement. This also highlights the importance of the stability of the ensemble model for this task.

\subsection{Reranker-level Ensemble Analysis}
\label{reranker level ensemble analysis}

In this section, we investigate the impact of reranker-level ensemble on RAG performance. The experimental setup is as follows: for each query, we retrieve ten relevant documents and use a base model to rank them, selecting the top five based on relevance.

Since the model may introduce hallucinations during ranking, we perform a post-check on the ranked list. Specifically, we remove duplicate document identifiers and filter out IDs outside the 1–10 range. For any missing identifiers after cleaning, we pad the sequence at the end to maintain a consistent length.

\begin{figure}[t]
    \centering
    \includegraphics[width=\linewidth]{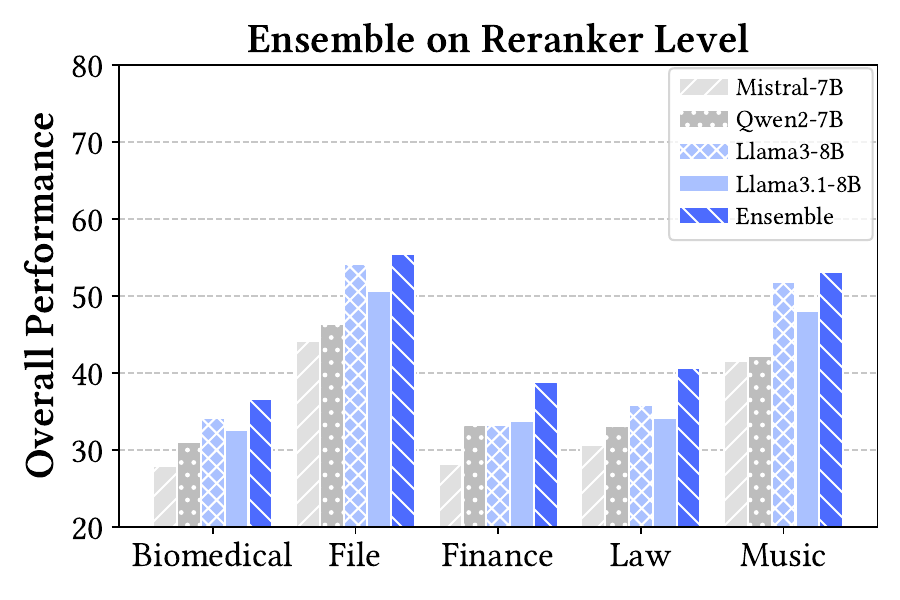}
    \caption{F1 Score for reranker level's ensemble on MS MARCO.}
    \label{fig:reranker}
\end{figure}

For the base model experiment, the same base model directly performs RAG over the top five ranked documents to generate an answer. For the ensemble experiment, we provide the ensemble model with three sets of top-ranked documents, each produced by a different base model. We explicitly indicate that each set is ordered by descending relevance, guiding the model to generate a final answer based on all fifteen documents. Note that the ensemble model receives only the documents as input, without any intermediate answers. The output behavior of the ensemble model is summarized in the figure~\ref{fig:reranker}.

Experimental results show that ensemble at the reranker level remains effective. The aggregated performance consistently surpasses that of standard RAG with reranking. Additionally, the model demonstrates strong robustness in ensemble. Since different base models may rank the same document differently, the aggregated input may contain conflicting relevance signals. Nevertheless, the model is still able to produce accurate final answers, suggesting a degree of self-discrimination and resistance to noise during ensemble.

\section{Conclusion}
In this paper, we perform a thorough analysis of the method for aggregating information from multiple RAG systems to derive comprehensive answers. This is the first detailed analysis of the ensemble method about RAG ensemble framework. We establish a mathematical model that provides a solid formulation of the ensemble process. 
In addition, we conduct a lot of experiments at both the pipeline level and the module level, fully demonstrating the broad adaptability, effectiveness, and stability of the RAG ensemble framework. Moreover, we have drawn some important conclusions. For example, we find that the RAG system ensemble framework exhibits a scaling-up phenomenon, and that the ensemble model has different preferences for tasks of varying difficulty levels. These conclusions are supported by a series of our experiments. We hope this paper serves as a reference for research on the RAG system ensemble and encourages further work to optimize RAG system performance.

\section*{GenAI Usage Disclosure}
In this paper, there is no use of GenAI tools whatsoever in any stage of the research.





\bibliographystyle{ACM-Reference-Format}
\bibliography{sample-base}

\appendix

\end{document}